\documentclass[runningheads]{llncs}
\pdfoutput=1
\usepackage{graphicx}
\usepackage{amsmath,amssymb} % define this before the line numbering.
\usepackage{color}
\usepackage[width=122mm,left=12mm,paperwidth=146mm,height=193mm,top=12mm,paperheight=217mm]{geometry}
\usepackage{array}
\usepackage{float}
\usepackage{subfig,caption}
\usepackage{lipsum}
\usepackage{comment}
\usepackage{epstopdf}
\usepackage{chngcntr}

\counterwithin*{section}{part}
\renewcommand{\arraystretch}{1.1}

\begin{document}

% \linenumbers
\pagestyle{headings}
\mainmatter
\title{Analyzing the Performance of Multilayer Neural Networks for Object Recognition} % Replace with your title

\titlerunning{Analyzing The Performance of Multilayer Neural Networks}

\authorrunning{Pulkit Agrawal, Ross Girshick, Jitendra Malik}

\author{Pulkit Agrawal, Ross Girshick, Jitendra Malik\\
\texttt{\small \{pulkitag,rbg,malik\}@eecs.berkeley.edu}}
\institute{University of California, Berkeley}

\maketitle

\begin{abstract}
In the last two years, convolutional neural networks (CNNs) have achieved an impressive suite of results on standard recognition datasets and tasks.
CNN-based features seem poised to quickly replace engineered representations, such as SIFT and HOG.
However, compared to SIFT and HOG, we understand much less about the nature of the features learned by large CNNs.
In this paper, we experimentally probe several aspects of CNN feature learning in an attempt to help practitioners gain useful, evidence-backed intuitions about how to apply CNNs to computer vision problems.
%Towards this end, our paper is an empirical study addressing the following four questions: 
%(1) Does a multilayer CNN learn ``grandmother'' cells or a distributed representation?
%(2) How does fine-tuning affect performance and is pre-training still useful when more data becomes available? 
%(3) Does longer pre-training time hurt generalization performance?
%(4) How important is a feature's spatial location and its activation magnitude?

\end{abstract}

\section{Introduction}
% Over the past two years we've seen a string of work
% Krizhevsky, Donahue, Girshick, and now many more demonstrating that
% convolutional neural networks are able to learn features that
% outperform traditional engineered features in a wide array of tasks.

Over the last two years, a sequence of results on benchmark visual recognition tasks has demonstrated that convolutional neural networks (CNNs) \cite{fukushima1980neocognitron,Lecun89,rumelhart86} will likely replace engineered features, such as SIFT \cite{Sift} and HOG \cite{Hog}, for a wide variety of problems.
This sequence started with the breakthrough ImageNet \cite{imagenet} classification results reported by Krizhevsky et al. \cite{Kriz}.
Soon after, Donahue et al. \cite{Decaf} showed that the same network, trained for ImageNet classification, was an effective blackbox feature extractor.
Using CNN features, they reported state-of-the-art results on several standard image classification datasets.
At the same time, Girshick et al. \cite{Rcnn} showed how the network could be applied to object detection.
Their system, called R-CNN, classifies object proposals generated by a bottom-up grouping mechanism (e.g., selective search \cite{UijlingsIJCV2013}).
Since detection training data is limited, they proposed a transfer learning strategy in which the CNN is first pre-trained, with supervision, for ImageNet classification and then fine-tuned on the small PASCAL detection dataset \cite{Pascal}.
Since this initial set of results, several other papers have reported similar findings on a wider range of tasks (see, for example, the outcomes reported by Razavian et al. in \cite{astounding}).

Feature transforms such as SIFT and HOG afford an intuitive interpretation as histograms of oriented edge filter responses arranged in spatial blocks.
However, we have little understanding of what visual features the different layers of a CNN encode.
Given that rich feature hierarchies provided by CNNs are likely to emerge as the prominent feature extractor for computer vision models over the next few years, we believe that developing such an understanding is an interesting scientific pursuit and an essential exercise that will help guide the design of computer vision methods that use CNNs.
Therefore, in this paper we study several aspects of CNNs through an empirical lens.

% Therefore in this paper, we study several aspects of CNNs
% - What is the nature of the feature representation? (grandmother vs. distributed)
% - How does fine-tuning affect performance?
% - How does pre-training time affect generalization?
% - Untangling feature magnitude and location

\subsection{Summary of findings}

\subsubsection{Effects of fine-tuning and pre-training.} 
Girshick et al. \cite{Rcnn} showed that supervised pre-training and fine-tuning are effective when training data is scarce.
However, they did not investigate what happens when training data becomes more abundant.
We show that it is possible to get good performance when training R-CNN from a random initialization (i.e., without ImageNet supervised pre-training) with a reasonably modest amount of detection training data (37k ground truth bounding boxes).
However, we also show that in this data regime, supervised pre-training is still beneficial and leads to a large improvement in detection performance.
We show similar results for image classification, as well.

\subsubsection{ImageNet pre-training does not overfit.}
One concern when using supervised pre-training is that achieving a better model fit to ImageNet, for example, might lead to higher generalization error when applying the learned features to another dataset and task.
If this is the case, then some form of regularization during pre-training, such as early stopping, would be beneficial.
We show the surprising result that pre-training for longer yields better results, with diminishing returns, but does \emph{not} increase generalization error.
This implies that fitting the CNN to ImageNet induces a general and portable feature representation.
Moreover, the learning process is well behaved and does not require ad hoc regularization in the form of early stopping.

\subsubsection{Grandmother cells and distributed codes.}
%Recent work on visualizing features from deep networks (e.g., \cite{GoogleCat,DeConv}) suggests that the features might consist mainly of ``grandmother'' cells \cite{Barlow,Grandmother}.
%In other words, that the CNN learns features that fire only when a specific semantic class, such as \emph{cat}, is present.
We do not have a good understanding of mid-level feature representations in multilayer networks.
Recent work on feature visualization, (e.g., \cite{GoogleCat,DeConv}) suggests that such networks might consist mainly of ``grandmother'' cells \cite{Barlow,Grandmother}.
%Our analysis shows that the representation is more subtle.
Our analysis shows that the representation in intermediate layers is more subtle.
There are a small number of grandmother-cell-like features, but most of the feature code is distributed and several features must fire in concert to effectively discriminate between classes.

\subsubsection{Importance of feature location and magnitude.}
Our final set of experiments investigates what role a feature's spatial location and magnitude plays in image classification and object detection.
Matching intuition, we find that spatial location is critical for object detection, but matters little for image classification.
More surprisingly, we find that feature magnitude is largely unimportant.
For example, binarizing features (at a threshold of 0) barely degrades performance.
This shows that sparse binary features, which are useful for large-scale image retrieval \cite{gong2011iterative,weiss2009spectral}, come ``for free'' from the CNN's representation.

\section{Experimental setup}
\label{sec:train}

\subsection{Datasets and tasks}
In this paper, we report experimental results using several standard datasets and tasks, which we summarize here.

\subsubsection{Image classification.} For the task of image classification we consider two datasets, the first of which is PASCAL VOC 2007 \cite{Pascal}.
We refer to this dataset and task by ``PASCAL-CLS''.
Results on PASCAL-CLS are reported using the standard average precision (AP) and mean average precision (mAP) metrics.

PASCAL-CLS is fairly small-scale with only 5k images for training, 5k images for testing, and 20 object classes.
Therefore, we also consider the medium-scale SUN dataset \cite{sun}, which has around 108k images and 397 classes.
We refer to experiments on SUN by ``SUN-CLS''.
In these experiments, we use a non-standard train-test split since it was computationally infeasible to run all of our experiments on the 10 standard subsets proposed by \cite{sun}. 
Instead, we randomly split the dataset into three parts (train, val, and test) using 50\%, 10\% and 40\% of the data, respectively. 
The distribution of classes was uniform across all the three sets.
We emphasize that results on these splits are only used to support investigations into properties of CNNs and not for comparing against other scene-classification methods in the literature.
For SUN-CLS, we report 1-of-397 classification accuracy averaged over all classes, which is the standard metric for this dataset\footnote{The version of this paper published at ECCV'14 contained an error in our description of the accuracy metric. That version used overall accuracy, instead of class-averaged accuracy. This version contains corrected numbers for SUN-CLS
to reflect the standard accuracy metric of class-averaged accuracy.}.
For select experiments we report the error bars in performance as mean $\pm$ standard deviation in accuracy over 3 runs (it was computationally infeasible to compute error bars for all experiments). For each run, a different random split of train, val, and test sets was used.  

\subsubsection{Object detection.} For the task of object detection we use PASCAL VOC 2007.
We train using the trainval set and test on the test set.
We refer to this dataset and task by ``PASCAL-DET''.
PASCAL-DET uses the same set of images as PASCAL-CLS.
We note that it is standard practice to use the 2007 version of PASCAL VOC for reporting results of ablation studies and hyperparameter sweeps.
We report performance on PASCAL-DET using the standard AP and mAP metrics.
In some of our experiments we use only the ground-truth PASCAL-DET bounding boxes, in which case we refer to the setup by ``PASCAL-DET-GT''.

In order to provide a larger detection training set for certain experiments, we also make use of the ``PASCAL-DET+DATA'' dataset, which we define as including VOC 2007 trainval union with VOC 2012 trainval.
The VOC 2007 test set is still used for evaluation.
This dataset contains approximately 37k labeled bounding boxes, which is roughly three times the number contained in PASCAL-DET.

\subsection{Network architecture and layer nomenclature}
\label{sub:net-arch}
All of our experiments use a single CNN architecture.
This architecture is the Caffe \cite{caffe} implementation of the network proposed by Krizhevsky et al. \cite{Kriz}.
The layers of the CNN are organized as follows.
The first two are subdivided into four sublayers each: convolution (conv), $\max(x,0)$ rectifying non-linear units (ReLUs), max pooling, and local response normalization (LRN). 
Layers 3 and 4 are composed of convolutional units followed by ReLUs.
Layer 5 consists of convolutional units, followed by ReLUs and max pooling.
The last two layers are fully connected (fc). 
When we refer to conv-1, conv-2, and conv-5 we mean the output of the max pooling units following the convolution and ReLU operations (also following LRN when applicable).\footnote{Note that this nomenclature differs slightly from \cite{Rcnn}.}
For layers conv-3, conv-4, fc-6, and fc-7 we mean the output of ReLU units.
%FT or FT-Net refers to a finetuned network whereas as FC-FT or FC-FT-Net refers to a network finetuned by setting the learning rate of the first 5 layers to zero. We use the terms CNNs and ConvNets interchangeably to refer to multilayer network architectures of the type proposed in \cite{Kriz}. Terms filter/unit are used interchangeably to refer to filters of the CNN and GT-BBOX/gt-bbox stands for Ground truth bounding boxes from the PASCAL-VOC-2007 detection challenge and mAP refers to mean average precision \cite{Pascal}.

%\subsection{Training Setup} 
%\label{sub:train-setup}
%Results for image and GT-BBOX classification were obtained by training linear SVMs on train-val sets of PASCAL-VOC-2007 \cite{Pascal} and tested on the test set. For detection we closely follow the RCNN setup described in \cite{Rcnn}. For SUN-397 \cite{sun} we used a non-standard train-test splits since it was infeasible to finetune CNNs for 10 standard subsets proposed by \cite{sun}. Instead, we randomly split the dataset into 3 parts namely train,val and test using 50\%,10\% and 40\% of the data. The distribution of classes was uniform across all the 3 sets. Results on these splits are only used to support investigations into properties of CNNs and not for comparing against other scene-classification methods.  
 
\subsection{Supervised pre-training and fine-tuning}
\label{sub:fine-train}
Training a large CNN on a small dataset often leads to catastrophic overfitting.
%The idea of supervised pre-training is to use a data-rich auxiliary dataset and task, such as ImageNet classification, to initialize the CNN parameters before training models on a small dataset.
%This procedure is called \emph{fine-tuning}.
The idea of supervised \emph{pre-training} is to use a data-rich auxiliary dataset and task, such as ImageNet classification, to initialize the CNN parameters. 
The CNN can then be used on the small dataset, directly, as a feature extractor (as in \cite{Decaf}).
Or, the network can be updated by continued training on the small dataset, a process called \emph{fine-tuning}.
%The R-CNN object detection work in \cite{Rcnn} demonstrated that fine-tuning is an effective strategy for training object detectors.
%\cite{Decaf} also demonstrated that such pre-training, even without fine-tuning, can lead to state-of-the-art results on various image classification tasks. 

%We employ the supervised pre-training, domain-specific fine-tuning paradigm used by R-CNN \cite{Rcnn} in many experiments.
%The idea of supervised pre-training is to use a data-rich auxiliary dataset and task, such as ImageNet classification, to initialize a CNN with large number of parameters before training on a small dataset. Such initialization procedure allows the network parameters to be modified to achieve good performance on a small dataset without overfitting the large network to it.
%This procedure allows the small dataset to be used while avoiding disastrously overfitting the large network to it.
%For experiments in which the network is pre-trained on ImageNet, stochastic gradient descent is run for 310000 iterations (66 epochs).

For fine-tuning, we follow the procedure described in \cite{Rcnn}.
First, we remove the CNN's classification layer, which was specific to the pre-training task and is not reusable.
Next, we append a new randomly initialized classification layer with the desired number of output units for the target task.
Finally, we run stochastic gradient descent (SGD) on the target loss function, starting from a learning rate set to $0.001$ ($1/10$-th the initial learning rate used for training the network for ImageNet classification). 
%This choice was made to prevent clobbering the CNN's initialization.
This choice was made to prevent clobbering the CNN's initialization to control overfitting. 
%This choice was made to ensure that the network parameters don't overfit CNN's initialization.
At every 20,000 iterations of fine-tuning we reduce the learning rate by a factor of 10.

%\subsection{Method of Entropy Computation}
%\label{sub:def-ent}
%We define the entropy of a filter with respect to a given set of image-label pairs in the following way. Each image, when passed through the convolutional neural network produces a $p \times p$ heatmap of filter responses. (e.g. p = 6, for a layer 5 filter). We vectorize this heatmap into a vector of scores of length $p^2$ and with each element of this vector we associate the class label of the image. Thus, for each image we have a score vector and a label vector of length $p^2$ each. Next, we concatenate score vectors and label vectors from N images into a giant score vector and a giant label vector  of size $Np^2$ each. Now for every score threshold we consider all the labels which have an associated score $\geq$ to this threshold score. The entropy of this set of labels is the entropy of the filter at this threshold. As this threshold changes, entropy traces out a curve which we call as the entropy curve.  

\section{The effects of fine-tuning and pre-training on CNN performance and parameters}
\label{sec:fine}
The results in \cite{Rcnn} (R-CNN) show that supervised pre-training for ImageNet classification, followed by fine-tuning for PASCAL object detection, leads to large gains over directly using features from the pre-trained network (without fine-tuning).
However, \cite{Rcnn} did not investigate three important aspects of fine-tuning: (1) What happens if we train the network ``from scratch'' (i.e., from a random initialization) on the detection data? (2) How does the amount of fine-tuning data change the picture? and (3) How does fine-tuning alter the network's parameters?
In this section, we explore these questions on object detection and image classification datasets.

\subsection{Effect of fine-tuning on CNN performance}
\label{sub:fine-performance}
\setlength{\tabcolsep}{2pt}
\begin{table}[t!]
\begin{center}
\caption{Comparing the performance of CNNs trained from scratch, pre-trained on ImageNet, and fine-tuned. PASCAL-DET+DATA includes additional data from VOC 2012 trainval. (Bounding-box regression was not used for detection results.)}
\vspace{0.3em}
\label{table:fine-effect}
\scalebox{0.95}{
\begin{tabular}{ccc|ccc|ccc}
\multicolumn{3}{c|}{\textbf{SUN-CLS}} & \multicolumn{3}{c|}{\textbf{PASCAL-DET}} & \multicolumn{3}{c}{\textbf{PASCAL-DET+DATA}} \\
scratch & pre-train & fine-tune  & scratch & pre-train & fine-tune & scratch & pre-train & fine-tune\\
\hline
$35.7 \pm 0.2$ & $48.4 \pm 0.1$ & $52.2 \pm 0.1$ & 40.7 & 45.5 & 54.1 & 52.3 & 45.5 & 59.2 \\ 
\end{tabular}}
\end{center}
\end{table}
\setlength{\tabcolsep}{1.4pt}
The main results of this section are presented in Table \ref{table:fine-effect}.
First, we focus on the detection experiments, which we implemented using the open source R-CNN code.
All results use features from layer fc-7.

Somewhat surprisingly, it's possible to get reasonable results (40.7\% mAP) when training the CNN from scratch using only the training data from VOC 2007 trainval (13k bounding box annotations).
However, this is still worse than using the pre-trained network, directly, without fine-tuning (45.5\%).
Even more surprising is that when the VOC 2007 trainval data is augmented with VOC 2012 data (an additional 25k bounding box annotations), we are able to achieve a mAP of 52.3\% from scratch.
This result is almost as good as the performance achieved by pre-training on ImageNet and then fine-tuning on VOC 2007 trainval (54.1\% mAP).
These results can be compared to the 30.5\% mAP obtained by DetectorNet \cite{szegedy2013deep}, a recent detection system based on the same network architecture, which was trained from scratch on VOC 2012 trainval.

Next, we ask if ImageNet pre-training is still useful in the PASCAL-DET +DATA setting?
Here we see that even though it's possible to get good performance when training from scratch, pre-training still helps considerably.
The final mAP when fine-tuning with the additional detection data is 59.2\%, which is 5 percentage points higher than the best result reported in \cite{Rcnn} (both without bounding-box regression).
This result suggests that R-CNN performance is not data saturated and that simply adding more detection training data without any other changes may substantially improve results.

We also present results for SUN image classification.
Here we observe a similar trend: reasonable performance is achievable when training from scratch, however initializing from ImageNet and then fine-tuning yields significantly better performance.

\subsection{Effect of fine-tuning on CNN parameters}
\label{sub:fine-entropy}
We have provided additional evidence that fine-tuning a discriminatively pre-trained network is very effective in terms of task performance.
Now we look inside the network to see how fine-tuning changes its parameters.

To do this, we define a way to measure the class selectivity of a set of filters.
Intuitively, we use the class-label entropy of a filter given its activations, above a threshold, on a set of images.
Since this measure is entropy-based, a low value indicates that a filter is highly class selective, while a large value indicates that a filter fires regardless of class.
The precise definition of this measure is given in the Appendix.

In order to summarize the class selectivity for a \emph{set} of filters, we sort them from the most selective to least selective and plot the average selectivity of the first $k$ filters while sweeping $k$ down the sorted list.
Figure \ref{fig:fine-entropy} shows the class selectivity for the sets of filters in layers 1 to 7 before and after fine-tuning (on VOC 2007 trainval).
Selectivity is measured using the ground truth boxes from PASCAL-DET-GT instead of a whole-image classification task to ensure that filter responses are a direct result of the presence of object categories of interest and not correlations with image background.

\begin{comment}
\begin{figure}[t!]
\centering
\subfloat{\includegraphics[height=6.5cm]{images/ent_hist.png}}
\caption{Distribution of AuE for different layers in Alex-Net and FT-Net. X-axis is the entropy and the Y-axis is the number of filters. Notice that the left tail for fc-6 and fc-7 becomes heavier after finetuning. This indicates that finetuning makes these filters more discriminative.}
\label{fig:fine-hist}
\end{figure}
\end{comment}

\begin{figure}[t!]
\centering
\subfloat{\includegraphics[scale=0.20]{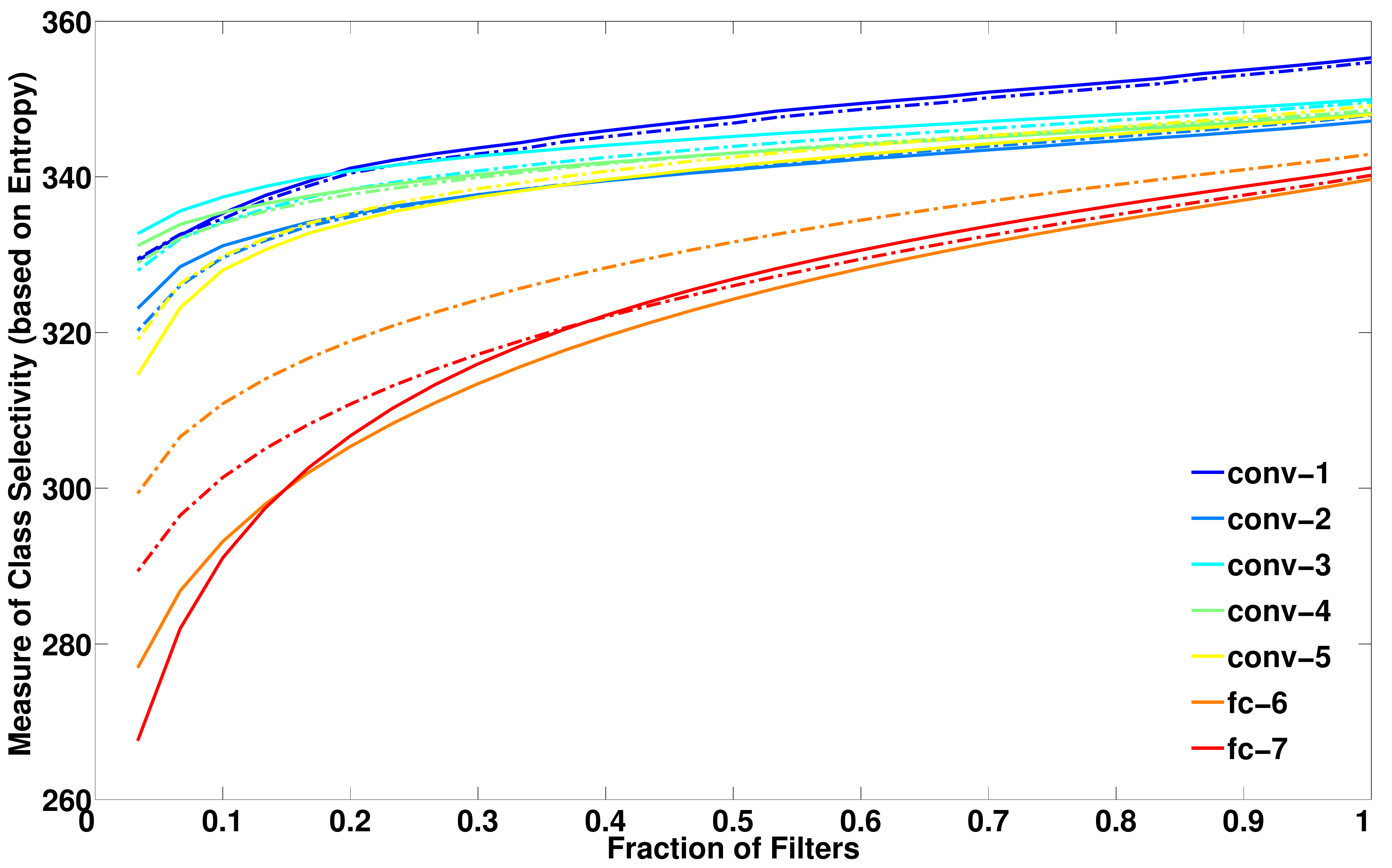}}
\caption{PASCAL object class selectivity plotted against the fraction of filters, for each layer, before fine-tuning (dash-dot line) and after fine-tuning (solid line). A lower value indicates greater class selectivity. Although layers become more discriminative as we go higher up in the network, fine-tuning on limited data (PASCAL-DET) only significantly affects the last two layers (fc-6 and fc-7).}
\label{fig:fine-entropy}
\end{figure}

\setlength{\tabcolsep}{2pt}
\begin{table}[t!]
\begin{center}
\caption{Comparison in performance when fine-tuning the entire network (ft) versus only fine-tuning the fully-connected layers (fc-ft).}
\label{table:fine-fc-ft}
\vspace{0.3em}
\scalebox{1}{
\begin{tabular}{cc|cc|cc}
\multicolumn{2}{c|}{\textbf{SUN-CLS}} & \multicolumn{2}{c|}{\textbf{PASCAL-DET}} & \multicolumn{2}{c}{\textbf{PASCAL-DET+DATA}} \\
ft & fc-ft  & ft & fc-ft & ft & fc-ft \\
\hline
$52.2 \pm 0.1$ & $51.6 \pm 0.1$ & 54.1 & 53.3 & 59.2 & 56.0 \\ 
\end{tabular}}
\end{center}
\end{table}
\setlength{\tabcolsep}{1.4pt}

Figure \ref{fig:fine-entropy} shows that class selectivity increases from layer 1 to 7 both with and without fine-tuning.
%This result is consistent with performance numbers reported in Table \ref{table:det-traj-classify}. 
It is interesting to note that entropy changes due to fine-tuning are only significant for layers 6 and 7. 
This observation indicates that fine-tuning only layers 6 and 7 may suffice for achieving good performance when fine-tuning data is limited. 
We tested this hypothesis on SUN-CLS and PASCAL-DET by comparing the performance of a fine-tuned network (ft) with a network which was fine-tuned by only updating the weights of fc-6 and fc-7 (fc-ft). 
These results, in Table \ref{table:fine-fc-ft}, show that with small amounts of data, fine-tuning amounts to ``rewiring'' the fully connected layers.
However, when more fine-tuning data is available (PASCAL-DET+DATA), there is still substantial benefit from fine-tuning all network parameters.

\subsection{Effect of pre-training on CNN parameters}
\label{sec:speed}
%\section{How does pre-training time affect generalization performance?}
%\label{sec:speed}
%Pre-Training is the process of initializing CNN parameters for a target application using images from a (generally larger) separate dataset. Features learned by a CNN pre-trained on Imagenet have been shown to generalize and achieve state of art results across multiple computer vision datasets (see section \ref{sec:fine} and \cite{Decaf}). Since, no single image dataset fully captures the variation in natural images, all datasets are biased (cite Alyosha). Consequently, it can be expected that excessive pre-training can cause the CNN to overfit on Imagenet and thus hurt generalization performance. 
There is no single image dataset that fully captures the variation in natural images. This means that all datasets, including ImageNet, are biased in some way. Thus, there is a possibility that pre-training may eventually cause the CNN to overfit and consequently hurt generalization performance \cite{torralba2011unbiased}. To understand if this happens, in the specific case of ImageNet pre-training, we investigated the effect of pre-training time on generalization performance both with and without fine-tuning. We find that pre-training for longer improves performance. This is surprising, as it shows that fitting more to ImageNet leads to better performance when moving to the other datasets that we evaluated.
%The other interesting thing we note is that within 50K iterations of pre-training, the performance is almost 90\% of the final performance achieved after nearly 300K iterations.

\setlength{\tabcolsep}{4pt}
\begin{table}[t!]
\begin{center}
\caption{Performance variation (\% mAP) on PASCAL-CLS as a function of pre-training iterations on ImageNet. The error bars  for all columns are similar to the one reported in the 305k column.}
\label{table:det-traj-classify}
\vspace{0.3em}
\begin{tabular}{lcccccccccc}
layer  & 5k & 15k & 25k & 35k & 50k & 95k & 105k & 195k & 205k & 305k \\
\hline
conv-1 & 23.0 & 24.3 & 24.4 & 24.5 & 24.3 & 24.8 & 24.7 & 24.4 & $24.4$  & $24.4 \pm 0.5$ \\
conv-2 & 33.7 & 40.4 & 40.9 & 41.8 & 42.7 & 43.2 & 44.0 & 45.0 & $45.1$  & $45.1 \pm 0.7$ \\
conv-3 & 34.2 & 46.8 & 47.0 & 48.2 & 48.6 & 49.4 & 51.6 & 50.7 & $50.9$  & $50.5 \pm 0.6$ \\
conv-4 & 33.5 & 49.0 & 48.7 & 50.2 & 50.7 & 51.6 & 54.1 & 54.3 & $54.4$  & $54.2 \pm 0.7$ \\
conv-5 & 33.0 & 53.4 & 55.0 & 56.8 & 57.3 & 59.2 & 63.5 & 64.9 & $65.5$  & $65.6 \pm 0.3 $ \\
fc-6   & 34.2 & 59.7 & 62.6 & 62.7 & 63.5 & 65.6 & 69.3 & 71.3 & $71.8$  & $72.1 \pm 0.3 $\\
fc-7   & 30.9 & 61.3 & 64.1 & 65.1 & 65.9 & 67.8 & 71.8 & 73.4 & $74.0$  & $74.3 \pm 0.3 $\\
\end{tabular}
\end{center}
\end{table}
\setlength{\tabcolsep}{1.4pt}

\begin{figure}[t!]
\centering
\subfloat[5k Iterations]{\includegraphics[scale=0.10]{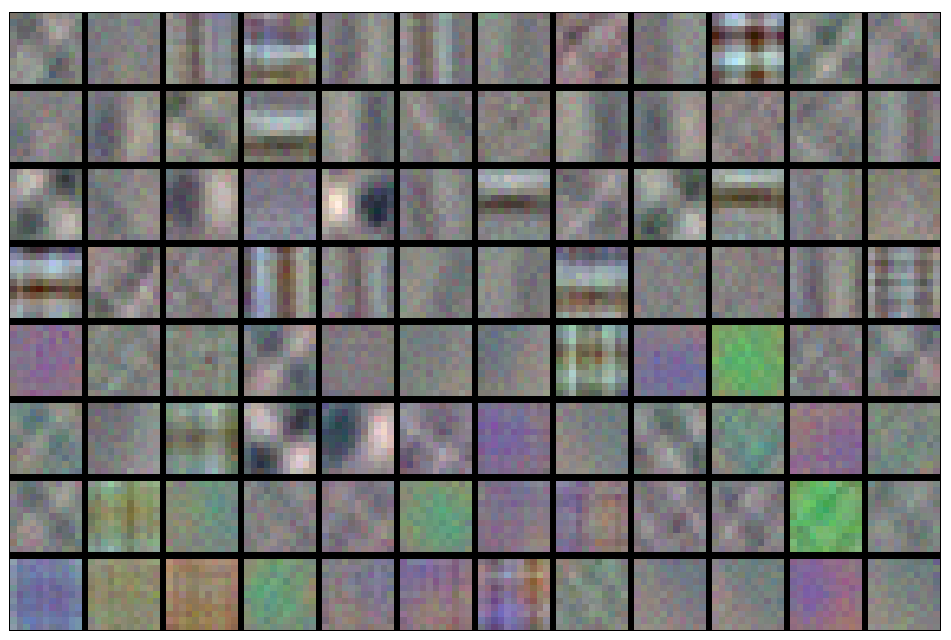}} \hspace{2mm}
\subfloat[15k Iterations]{\includegraphics[scale=0.10]{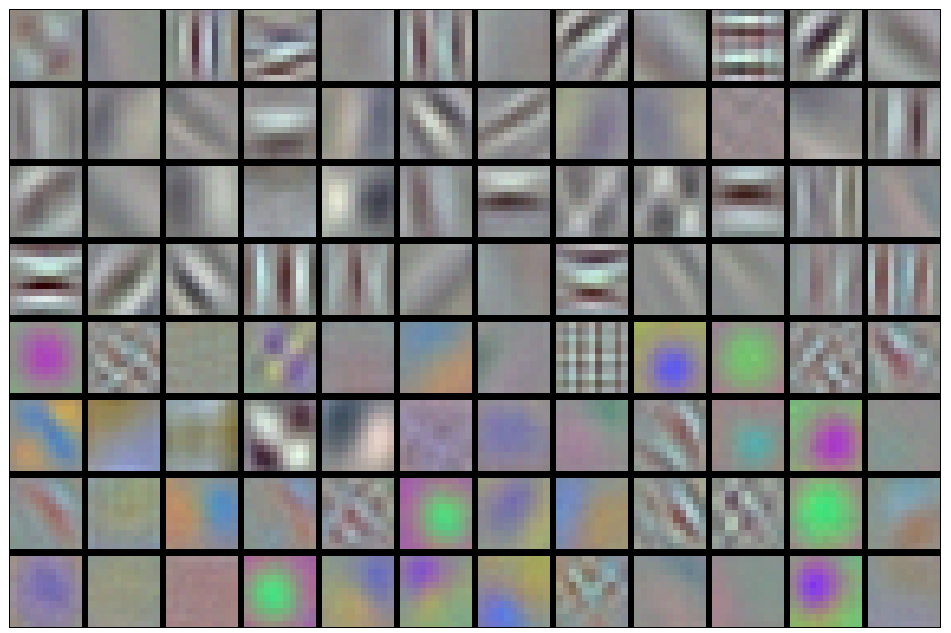}} \hspace{2mm}
\subfloat[305k Iterations]{\includegraphics[scale=0.10]{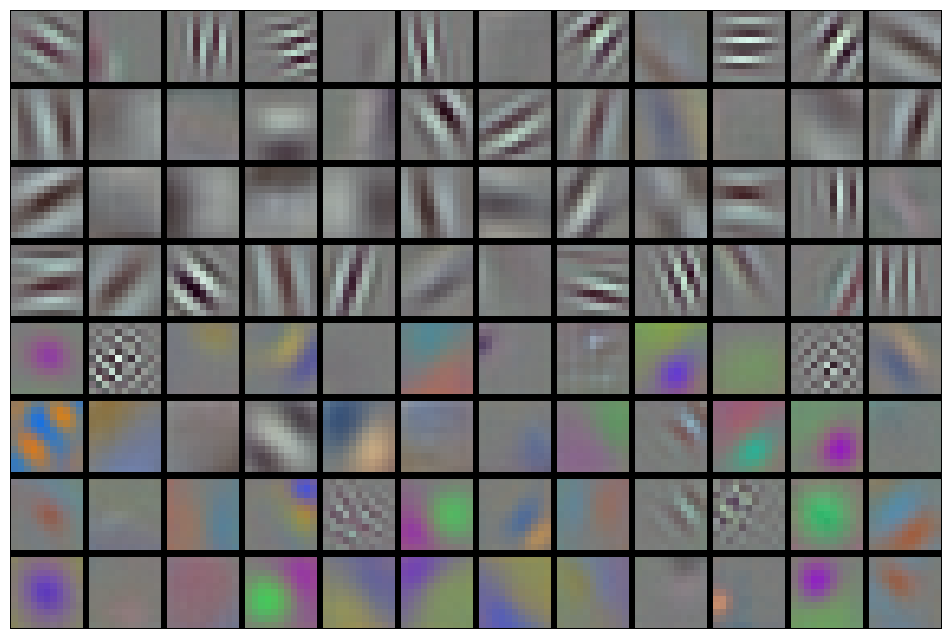}}
%\caption{(a), (b), and (c) show conv-1 filters after 5k, 15k and 305k iterations of training, respectively. Notice that after just 15k iterations these filters closely resemble their final state.}
\caption{Evolution of conv-1 filters with time. After just 15k iterations, these filters closely resemble their converged state.}
\begin{comment} Note: I have labeled 305k as 225k - I cannot get the same shape as of the 5k, 15k for 305k. The filters look very similar and visually indistinguishable from 305k. If I have time later, I will make them all uniform.
\end{comment}
\label{fig:conv1}
\end{figure}

We report performance on PASCAL-CLS as a function of pre-training time, without fine-tuning, in Table \ref{table:det-traj-classify}. Notice that more pre-training leads to better performance. By 15k and 50k iterations all layers are close to 80\% and  90\% of their final performance (5k iterations is only $\sim$1 epoch). This indicates that training required for generalization takes place quite quickly. Figure \ref{fig:conv1} shows conv-1 filters after 5k, 15k, and 305k iterations and reinforces this observation. Further, notice from Table \ref{table:det-traj-classify} that conv-1 trains first and the higher the layer is the more time it takes to converge. This suggests that a CNN, trained with backpropagation, converges in a layer-by-layer fashion.
Table \ref{table:det-trajectory} shows the interaction between varied amounts of pre-training time and fine-tuning on SUN-CLS and PASCAL-DET.
Here we also see that more pre-training prior to fine-tuning leads to better performance.

\setlength{\tabcolsep}{4pt}
\begin{table}[t!]
\begin{center}
\caption{Performance variation on SUN-CLS and PASCAL-DET using features from a CNN pre-trained for different numbers of iterations and fine-tuned for a fixed number of iterations (40k for SUN-CLS and 70k for PASCAL-DET)}
\label{table:det-trajectory}
\vspace{0.3em}
\scalebox{1.00}{
\begin{tabular}{l|c|c|c|c}
 & 50k & 105k & 205k & 305k \\
\hline
\textbf{SUN-CLS} & $48.5 \pm 0.1 $ & $50.0 \pm 0.2 $ & $51.8 \pm 0.3 $ & $51.9 \pm 0.3$ \\
\textbf{PASCAL-DET} & 50.2 & 52.6 & 55.3 & 55.4\footnotemark \\
\end{tabular}}
\end{center}
\end{table}
\setlength{\tabcolsep}{1.4pt} 

\footnotetext{A network pre-trained from scratch, which was different from the one used in Section \ref{sub:fine-performance}, was used to obtain these results. The difference in performance is not significant.}

\section{Are there grandmother cells in CNNs?}
\label{sec:grand-mother}
Neuroscientists have conjectured that cells in the human brain which only respond to very specific and complex visual stimuli (such as the face of one's grandmother) are involved in object recognition.
These neurons are often referred to as \emph{grandmother cells} (GMC) \cite{Barlow,Grandmother}. 
Proponents of artificial neural networks have shown great interest in reporting the presence of GMC-like filters for specific object classes in their networks (see, for example, the cat filter reported in \cite{GoogleCat}). 
The notion of GMC like features is also related to standard feature encodings for image classification.
Prior to the work of \cite{Kriz}, the dominant approaches for image and scene classification were based on either representing images as a bag of local descriptors (BoW), such as SIFT (e.g., \cite{lazebnik2006beyond}), or by first finding a set of mid-level patches \cite{Blocks,Mid1} and then encoding images in terms of them. 
The problem of finding good mid-level patches is often posed as a search for a set of high-recall discriminative templates. 
In this sense, mid-level patch discovery is the search for a set of GMC templates. 
The low-level BoW representation, in contrast, is a \emph{distributed code} in the sense that a single feature by itself is not discriminative, but a group of features taken together is.
This makes it interesting to investigate the nature of mid-level CNN features such as conv-5. 

%Recently, proponents of artificial neural networks have shown great interest in reporting the presence of GMC-like filters for specific object classes in their networks (see, for example, the cat filter reported in \cite{GoogleCat}). 
For understanding these feature representations in CNNs, \cite{Simonyan,DeConv} recently presented methods for finding locally optimal visual inputs for individual filters.
%These methods isolate individual inputs that activate a filter, but do not characterize the \emph{distribution} of images that cause an individual filter to fire above a certain threshold. \todo{make this argument clearer}.
However, these methods only find the best, or in some cases top-$k$, visual inputs that activate a filter, but do not characterize the \emph{distribution} of images that cause an individual filter to fire above a certain threshold. For example, if it is found that the top-10 visual inputs for a particular filter are cats, it remains unclear what is the response of the filter to other images of cats.
Thus, it is not possible to make claims about presence of GMC like filters for cat based on such analysis. 
A GMC filter for the cat class, is one that fires strongly on \emph{all} cats and nothing else.
%If, for example, we wish to see if the network contains a GMC filter for the \emph{cat} class, then we should look for a particular filter that fires strongly on all cats and nothing else.
This criteria can be expressed as a filter that has high \emph{precision} and high \emph{recall}.
That is, a GMC filter for class $C$ is a filter that has a high average precision (AP) when tasked with classifying inputs from class $C$ versus inputs from all other classes.

First, we address the question of finding GMC filters by computing the AP of individual filters (Section \ref{sub:class-specific-unit}). Next, we measure how distributed are the feature representations (Section \ref{sub:how-many}). For both experiments we use features from layer conv-5, which consists of responses of 256 filters in a $6\times 6$ spatial grid. Using max pooling, we collapse the spatial grid into a 256-D vector, so that for each filter we have a single response per image (in Section \ref{sub:imp-mag} we show that this transformation causes only a small drop in task performance).

%The output of conv-5 is the response of 256 filters in a $6\times 6$ spatial grid which we collapse into a 256-D vector using max-pooling (see section \ref{sub:imp-loc} for more details)

\setlength{\unitlength}{\linewidth}
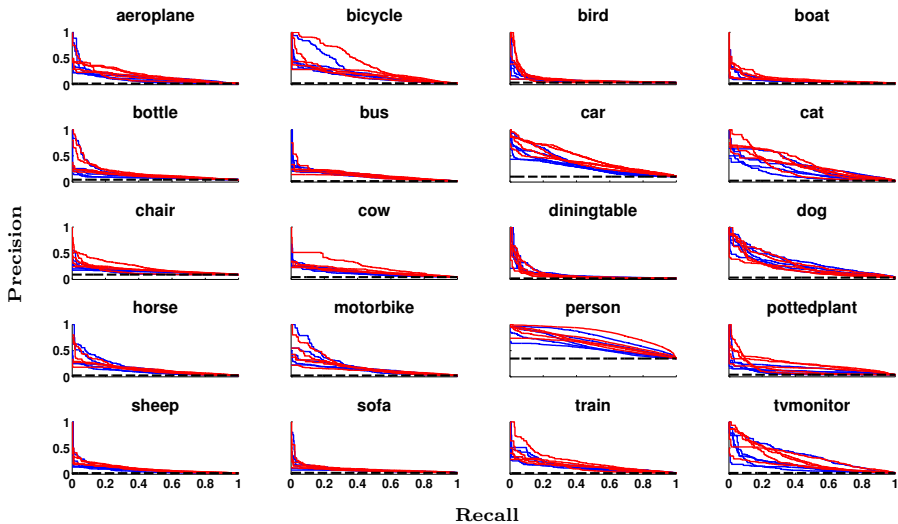
\begin{figure}[t!]
\centering
%\subfloat{\includegraphics[scale=0.20]{images/gtbbox_pascal_prerec_pool5units.png}}
\begin{picture}(0.04,0.3)(0,0)
\put(0.0,0.20){\rotatebox{90}{\scriptsize{\textbf{Precision}}}}
\end{picture}
\subfloat{\includegraphics[width=0.93\linewidth]{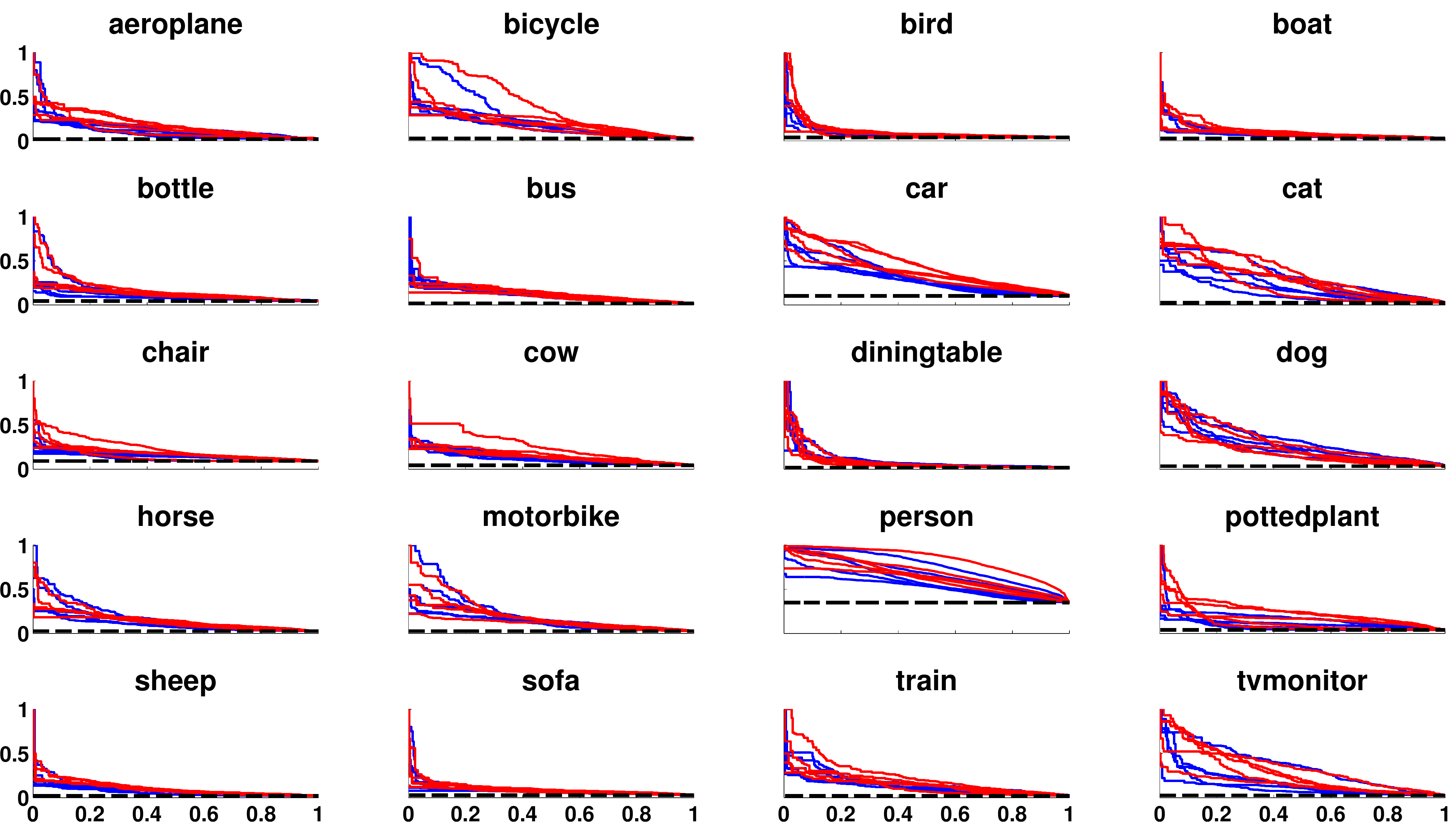}} \vspace{0.4mm}
\begin{picture}(1.0,0.01)(0,0)
\put(0.50,0.0){{\scriptsize{\textbf{Recall}}}}
\end{picture}
\caption{The precision-recall curves for the top five (based on AP) conv-5 filter responses on PASCAL-DET-GT. Curves in red and blue indicate AP for fine-tuned and pre-trained networks, respectively. The dashed black line is the performance of a random filter. For most classes, precision drops significantly even at modest recall values. There are GMC filters for classes such as bicycle, person, car, cat.}
\label{fig:ap}
\end{figure}

\subsection{Finding Grandmother Cells}
\label{sub:class-specific-unit}
For each filter, its AP value is calculated for classifying images using class labels and filter responses to object bounding boxes from PASCAL-DET-GT. Then, for each class we sorted filters in decreasing order of their APs. If GMC filters for this class exist, they should be the top ranked filters in this sorted list. The precision-recall curves for the top-five conv-5 filters are shown in Figure \ref{fig:ap}. We find that GMC-like filters exist for only for a few classes, such as bicycle, person, cars, and cats.

%\begin{figure}[t!]
%\centering
%\includegraphics[scale=0.20]{images/prob_sel_dims_top5.png}
%\caption{This plot shows the precision curve for the top 5 most selective filters taken from Alex-Net (Blue) and FT-Net(Red) for all PASCAL classes. Y-axis is the precision and X-axis is number of examples.}
%\label{fig:prob-sel}
%\end{figure}

\subsection{How distributed are the feature representations?}
\label{sub:how-many}

In addition to visualizing the AP curves of individual filters, we measured the number of filters required to recognize objects of a particular class. 
%Feature selection was performed to construct nested subsets of filters, ranging from a single filter to all filters, using the greedy strategy described in Figure \ref{fig:sel-strategy}.\footnote{Filter subsets of size [1, 2, 3, 5, 10, 15, 20, 25, 30, 35, 40, 45, 50, 80, 100, 128, 256] were used.}
Feature selection was performed to construct nested subsets of filters, ranging from a single filter to all filters, using the following greedy strategy. First, separate linear SVMs were trained to classify object bounding boxes from PASCAL-DET-GT using conv-5 responses. 
For a given class, the 256 dimensions of the learnt weight vector ($w$) is in direct correspondence with the 256 conv-5 filters. We used the magnitude of the $i$-th dimension of $w$ to rank the importance of the $i$-th conv-5 filter for discriminating instances of this class. 
Next, all filters were sorted using these magnitude values. Each subset of filters was constructed by taking the top-$k$ filters from this list.\footnote{We used values of $k \in \{1, 2, 3, 5, 10, 15, 20, 25, 30, 35, 40, 45, 50, 80, 100, 128, 256\}$.} For each subset, a linear SVM was trained using only the responses of filters in that subset for classifying the class under consideration.

\setlength{\unitlength}{\linewidth}
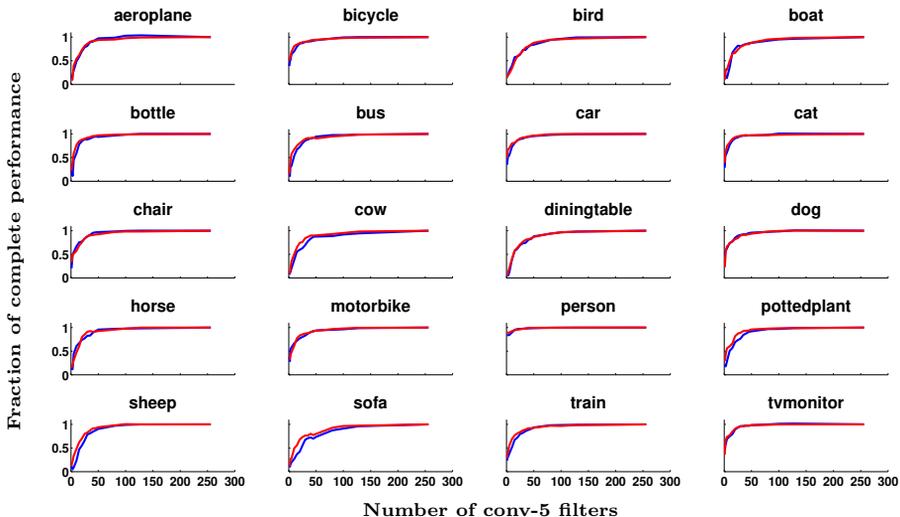
\begin{figure}[t!]
\centering
\begin{picture}(0.04,0.3)(0,0)
\put(0.0,0.06){\rotatebox{90}{\scriptsize{\textbf{Fraction of complete performance}}}}
\end{picture}
\includegraphics[width=0.93\linewidth]{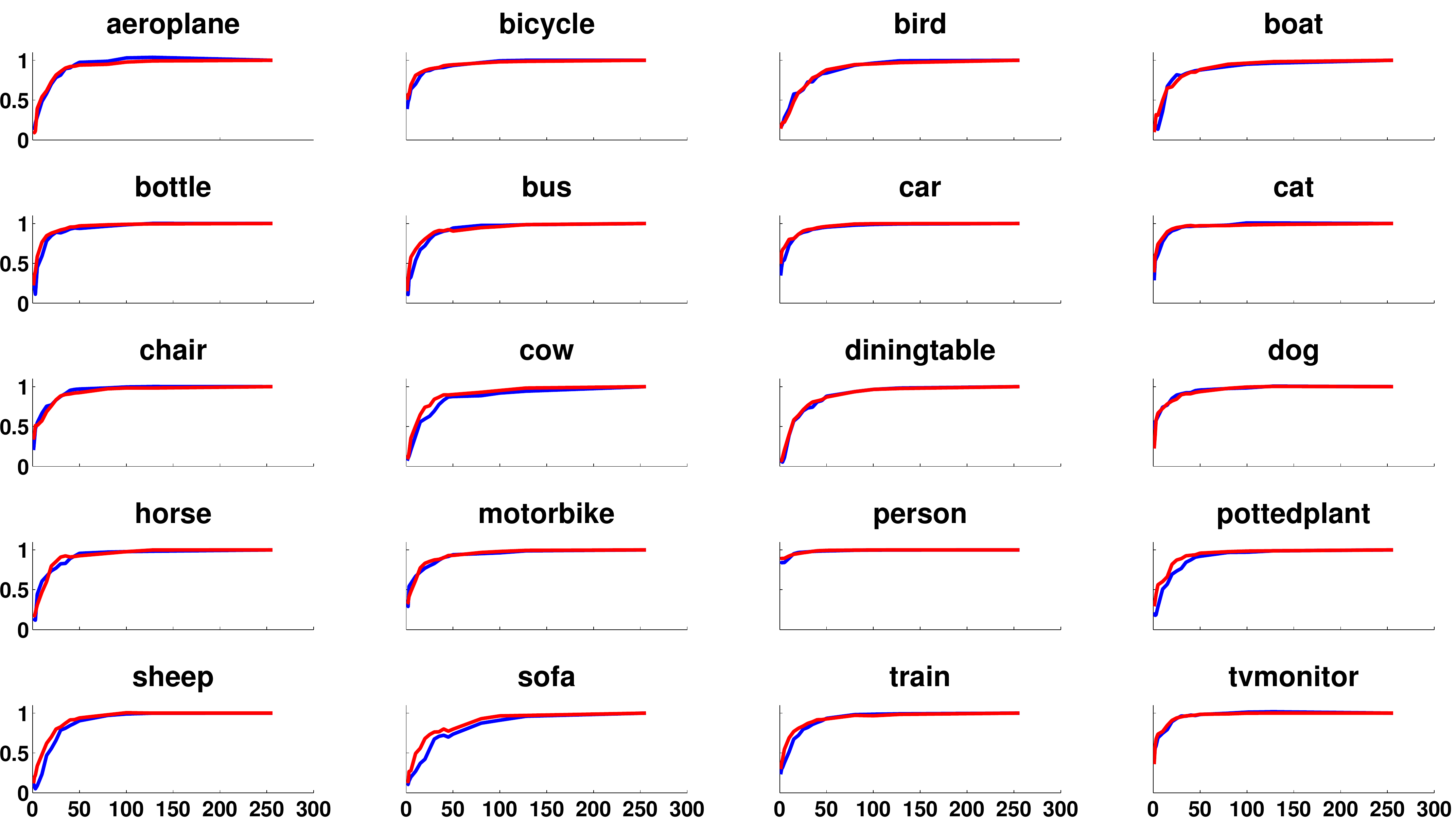}
\begin{picture}(1.0,0.01)(0,0)
\put(0.40,0.0){{\scriptsize{\textbf{Number of conv-5 filters}}}}
\end{picture}
\caption{The fraction of complete performance on PASCAL-DET-GT achieved by conv-5 filter subsets of different sizes. Complete performance is the AP computed by considering responses of all the filters. Notice, that for a few classes such as person and bicycle only a few filters are required, but for most classes substantially more filters are needed, indicating a distributed code.}
\label{fig:svm-sel-dims}
\end{figure}  

\setlength{\tabcolsep}{1.1pt}
\begin{table}[t!]
\renewcommand{\arraystretch}{1.2}
\begin{center}
\caption{Number of filters required to achieve 50\% or 90\% of the complete performance on PASCAL-DET-GT using a CNN pre-trained on ImageNet and fine-tuned for PASCAL-DET using conv-5 features.}
\label{table:num-fil}
\resizebox{\linewidth}{!}{
\begin{tabular}{l|c||cccccccccccccccccccc}
\noalign{\smallskip}
& perf. & aero & bike & bird & boat & bottle & bus & car & cat & chair & cow & table & dog & horse & mbike & person & plant & sheep & sofa & train & tv \\
\hline
pre-train & 50\% & 15 & 3 & 15 & 15 & 10 & 10 & 3 & 2 & 5 & 15 & 15 & 2 & 10 & 3 & 1 & 10 & 20 & 25 & 10 & 2 \\ 
fine-tune & 50\% & 10 & 1 & 20 & 15 & 5 & 5 & 2 & 2 & 3 & 10 & 15 & 3 & 15 & 10 & 1 & 5 & 15 & 15 & 5 & 2 \\
\hline
pre-train & 90\% & 40 & 35 & 80 & 80 & 35 & 40 & 30 & 20 & 35 & 100 & 80 & 30 & 45 & 40 & 15 & 45 & 50 & 100 & 45 & 25 \\
fine-tune & 90\% & 35 & 30 & 80 & 80 & 30 & 35 & 25 & 20 & 35 & 50 & 80 & 35 & 30 & 40 & 10 & 35 & 40 & 80 & 40 & 20 \\
\end{tabular}
}
\end{center}
\end{table}
\setlength{\tabcolsep}{1.4pt}
  
The variation in performance with the number of filters is shown in Figure \ref{fig:svm-sel-dims}.
Table \ref{table:num-fil} lists the number of filters required to achieve 50\% and 90\% of the complete performance. For classes such as persons, cars, and cats relatively few filters are required, but for most classes around 30 to 40 filters are required to achieve at least 90\% of the full performance. This indicates that the conv-5 feature representation is distributed and there are GMC-like filters for only a few classes.
Results using layer fc-7 are presented in the the supplementary material.
We also find that after fine-tuning, slightly fewer filters are required to achieve performance levels similar to a pre-trained network. 

\begin{figure}[t!]
\centering
\includegraphics[width=1.0\linewidth]{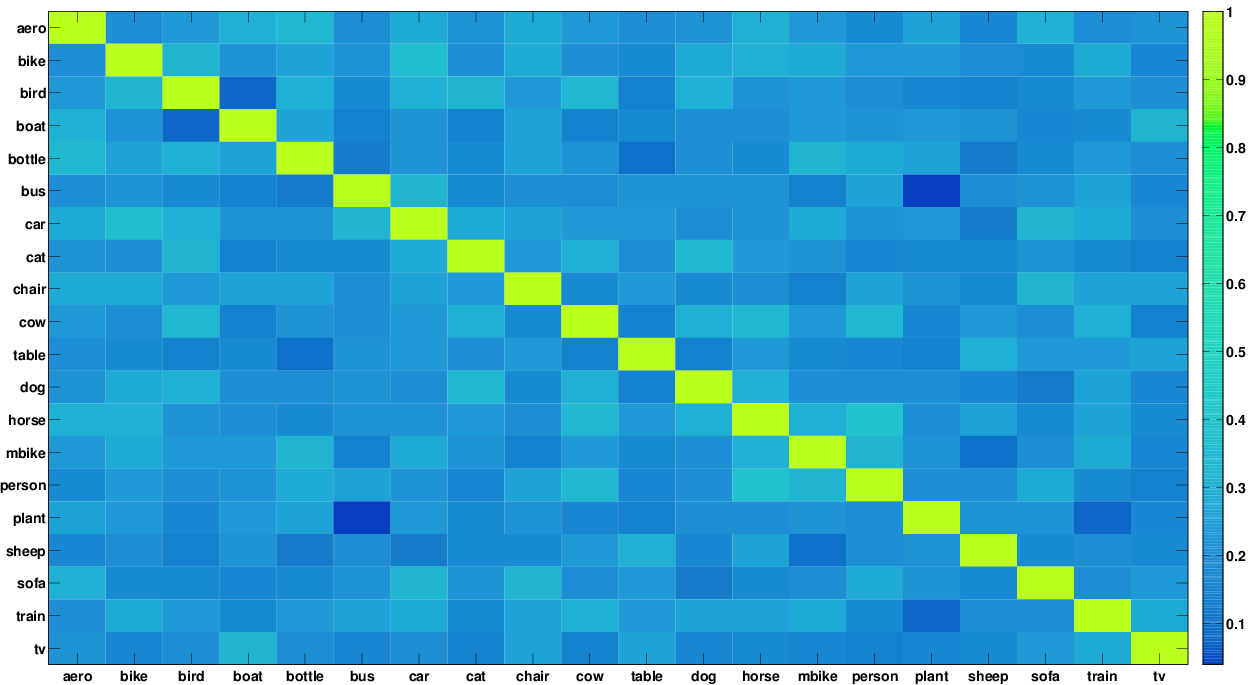}
\caption{The set overlap between the 50 most discriminative conv-5 filters for each class determined using PASCAL-DET-GT.
Entry $(i, j)$ of the matrix is the fraction of top-50 filters class $i$ has in common with class $j$ (Section \ref{sub:how-many}). Chance is 0.195. There is little overlap, but related classes are more likely to share filters.}
\label{fig:overlap}
\end{figure}

Next, we estimated the extent of overlap between the filters used for discriminating between different classes.
For each class $i$, we selected the 50 most discriminative filters (out of 256) and stored the selected filter indices in the set $S_i$.
The extent of overlap between class $i$ and $j$ was evaluated by $|S_i \cap S_j| / N$,
where $N = |S_i| = |S_j| = 50$. The results are visualized in Figure \ref{fig:overlap}. It can be seen that different classes use different subsets of conv-5 filters and there is little overlap between classes. This further indicates that intermediate representations in the CNN are distributed.
%and different sets of are used to distinguish between different classes. 

\section{Untangling feature magnitude and location}
\label{sec-where-info}
The convolutional layers preserve the coarse spatial layout of the network's input.
By layer conv-5, the original $227 \times 227$ input image has been progressively downsampled to $6 \times 6$.
This feature map is also sparse due to the $\max(x, 0)$ non-linearities used in the network (conv-5 is roughly 27\% non-zero; sparsity statistics for all layers are given in Table \ref{table:sparse}).
Thus, a convolutional layer encodes information in terms of (1) which filters have non-zero responses, (2) the magnitudes of those responses, and (3) their spatial layout.
In this section, we experimentally analyze the role of filter response magnitude and spatial location by looking at ablation studies on classification and detection tasks.

\begin{figure}[t!]
\centering
\includegraphics[height=6.5cm]{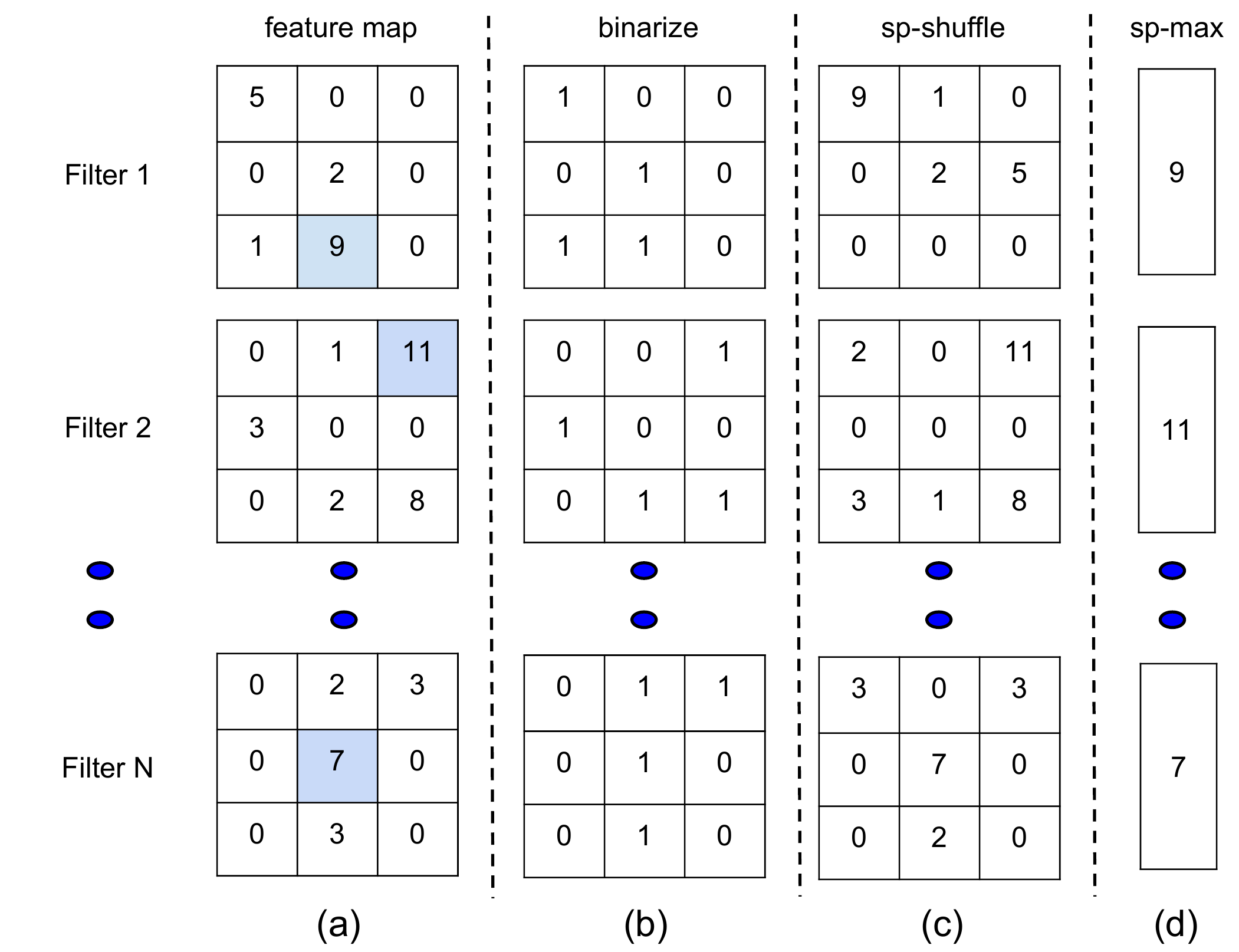}
\caption{Illustrations of ablations of feature activation spatial and magnitude information.
See Sections \ref{sub:imp-mag} and \ref{sub:imp-loc} for details.}
\label{fig:features}
\end{figure}

\subsection{How important is filter response magnitude?}
\label{sub:imp-mag}
\setlength{\tabcolsep}{4pt}
\begin{table}[t!]
\begin{center}
%\caption{Percentage non-zeros (sparsity) in filter responses of various layers of CNN.}
\caption{Percentage non-zeros (sparsity) in filter responses of CNN.}
\label{table:sparse}
\scalebox{1}{
\begin{tabular}{c|c|c|c|c|c|c}
conv-1 & conv-2 & conv-3 & conv-4 & conv-5 & fc-6 & fc-7 \\
\hline
$87.5 \pm 4.4$ & $44.5 \pm 4.4$ & $31.8 \pm 2.4$ & $32.0 \pm 2.7$ & $27.7 \pm 5.0$ & $16.1 \pm 3.0$ & $21.6 \pm 4.9$ \\
\end{tabular}}
\end{center}
\end{table}
\setlength{\tabcolsep}{1.4pt}

We can asses the importance of magnitude by setting each filter response $x$ to 1 if $x > 0$ and to $0$ otherwise. This binarization is performed prior to using the responses as features in a linear classifier and leads to loss of information contained in the magnitude of response while still retaining information about which filters fired and where they fired. 
In Tables \ref{table:class-ablation} and \ref{table:det-ablation} we show that binarization leads to a negligible performance drop for both classification and detection. 

For the fully-connected layers (fc-6 and fc-7) PASCAL-CLS performance is nearly identical before and after binarization.
This is a non-trivial property since transforming traditional computer vision features into short (or sparse) binary codes is an active research area. Such codes are important for practical applications in large-scale image retrieval and mobile image analysis \cite{gong2011iterative,weiss2009spectral}. Here we observe that sparse binary codes come essentially ``for free'' when using the representations learned in the fully-connected layers.

\subsection{How important is response location?}
\label{sub:imp-loc}
Now we remove spatial information from filter responses while retaining information about their magnitudes. We consider two methods for ablating spatial information from features computed by the convolutional layers (the fully-connected layers do not contain \emph{explicit} spatial information).

The first method (``sp-max'') simply collapses the $p \times p$ spatial map into a single value per feature channel by max pooling. The second method (``sp-shuffle'') retains the original distribution of feature activation values, but scrambles spatial correlations between columns of feature channels. To perform sp-shuffle, we permute the spatial locations in the $p \times p$ spatial map. This permutation is performed independently for each network input (i.e., different inputs undergo different permutations). Columns of filter responses in the same location move together, which preserves correlations between features within each (shuffled) spatial location. These transformations are illustrated in Figure \ref{fig:features}.

For image classification, damaging spatial information leads to a large difference in performance between original and spatially-ablated conv-1 features, but with a gradually decreasing difference for higher layers (Table \ref{table:class-ablation}). 
In fact, the performance of conv-5 after sp-max is close to the original performance. 
This indicates that a lot of information important for classification is encoded in the activation of the filters and not necessarily in the spatial pattern of their activations.
Note, this observation is not an artifact of small number of classes in PASCAL-CLS. On ImageNet validation data, conv-5 features and conv-5 after sp-max result into accuracy of 43.2 and 41.5 respectively. 
However, for detection sp-max leads to a large drop in performance. 
This may not be surprising since detection requires spatial information for precise localization.

\setlength{\tabcolsep}{4pt}
\begin{table}[t!]
\begin{center}
%\caption{Location and magnitude ablation study on PASCAL-CLS.}
\caption{Effect of location and magnitude feature ablations on PASCAL-CLS.}
\label{table:class-ablation}
\begin{tabular}{lccccc}
\noalign{\smallskip}
layer & no ablation (mAP) & binarize (mAP) & sp-shuffle (mAP) & sp-max (mAP) \\
\noalign{\smallskip}
\hline
\noalign{\smallskip}
conv-1 & $25.1 \pm 0.5$ & $17.7 \pm 0.2$ & $15.1 \pm 0.3$ & $25.4 \pm 0.5$  \\ 
conv-2 & $45.3 \pm 0.5$ & $43.0 \pm 0.6$ & $32.9 \pm 0.7$ & $40.1 \pm 0.3$  \\ 
conv-3 & $50.7 \pm 0.6$ & $47.2 \pm 0.6$ & $41.0 \pm 0.8$ & $54.1 \pm 0.5$  \\
conv-4 & $54.5 \pm 0.7$ & $51.5 \pm 0.7$ & $45.2 \pm 0.8$ & $57.0 \pm 0.5$  \\
conv-5 & $65.6 \pm 0.6$ & $60.8 \pm 0.7$ & $59.5 \pm 0.4$ & $62.5 \pm 0.6$  \\
fc-6   & $71.7 \pm 0.3$ & $71.5 \pm 0.4$ &  -             &  -   \\
fc-7   & $74.1 \pm 0.3$ & $73.7 \pm 0.4$ &  -             &  -   \\
\end{tabular}
\end{center}
\end{table}
\setlength{\tabcolsep}{1.4pt}

\setlength{\tabcolsep}{4pt}
\begin{table}[t!]
\begin{center}
\caption{Effect of location and magnitude feature ablations on PASCAL-DET.}
\label{table:det-ablation}
\scalebox{1.0}{
\begin{tabular}{l|c|c|c}
& no ablation (mAP) & binarize (mAP) & sp-max (mAP) \\
\hline
conv-5 & 47.6 & 45.7 & 25.4 
\end{tabular}}
\end{center}
\end{table}
\setlength{\tabcolsep}{1.4pt}

\begin{comment}
\setlength{\tabcolsep}{1pt}
\begin{table}[t!]
\begin{center}
\caption{Ablation study on PASCAL-DET using conv-5 features. Feature binarization leads to negligible drop in performance whereas as sp-max causes a large drop.}
\label{table:det-ablation}
\scalebox{0.70}{
\begin{tabular}{l|cccccccccccccccccccc||c}
\noalign{\smallskip}
& aero & bike & bird & boat & bottle & bus & car & cat & chair & cow & table & dog & horse & mbike & person & plant & sheep & sofa & train & tv & mAP \\
\noalign{\smallskip}
\hline
conv-5  & 57.8 & 63.9 & 38.8 & 28.0 & 29.0 & 54.8 & 66.9 & 51.3 & 30.5 & 52.1 & 45.2 & 43.2 & 57.3 & 58.8 & 46.0 & 27.2 & 51.2 & 39.3 & 53.3 & 56.6 & 47.6 \\
binarize & 57.9 & 61.3 & 32.6 & 24.7 & 27.5 & 55.0 & 64.7 & 49.8 & 25.3 & 47.4 & 44.5 & 40.3 & 54.6 & 56.4 & 43.6 & 27.1 & 48.4 & 41.6 & 54.3 & 57.6 & 45.7 \\
sp-max & 35.0 & 38.7 & 17.3 & 16.9 & 13.9 & 38.4 & 45.6 & 29.2 & 11.0 & 20.2 & 21.0 & 23.5 & 27.2 & 37.0 & 20.5 & 7.0 & 30.3 & 13.4 & 28.3 & 32.9 & 25.4 \\
\noalign{\smallskip}
\end{tabular}}
\end{center}
\end{table}
\setlength{\tabcolsep}{1.4pt}
\end{comment}

\section{Conclusion}
To help researchers better understand CNNs, we investigated pre-training and fine-tuning behavior on three classification and detection datasets.
We found that the large CNN used in this work can be trained from scratch using a surprisingly modest amount of data.
But, importantly, pre-training significantly improves performance and pre-training for longer is better.
We also found that some of the learnt CNN features are grandmother-cell-like, but for the most part they form a distributed code.
This supports the recent set of empirical results showing that these features generalize well to other datasets and tasks.

{\small
\smallskip
\noindent \textbf{Acknowledgments.}
This work was supported by ONR MURI N000141010933.
Pulkit Agrawal is partially supported by a Fulbright Science and Technology fellowship.
We thank NVIDIA for GPU donations.
We thank Bharath Hariharan, Saurabh Gupta and Jo\~{a}o Carreira for helpful suggestions.} \\

{\small
\smallskip
\noindent \textbf{Citing this paper.} 
Please cite the paper as: }\\ \\
\scriptsize{
@inproceedings\{agrawal14analyzing, \\
  Author    = \{Pulkit Agrawal and 
               Ross Girshick and 
               Jitendra Malik\}, \\                          
  Title     = \{Analyzing the Performance of Multilayer
               Neural Networks for Object Recognition\}, \\
  Booktitle = \{Proceedings of the European
               Conference on Computer Vision ({ECCV})\}, \\
  Year      = \{2014 \} \} \\
}

\normalsize{}

\section*{Appendix: estimating a filter's discriminative capacity}
\label{sub:app-entropy}
To measure the discriminative capacity of a filter, we collect filter responses from a set of $N$ images.
Each image, when passed through the CNN produces a $p \times p$ heat map of scores for each filter in a given layer (e.g., $p = 6$ for a conv-5 filter and $p = 1$ for an fc-6 filter).
This heat map is vectorized into a vector of scores of length $p^2$. With each element of this vector we associate the image's class label. 
Thus, for every image we have a score vector and a label vector of length $p^2$ each.
Next, the score vectors from all $N$ images are concatenated into an $Np^2$-length score vector.
The same is done for the label vectors.

Now, for a given score threshold $\tau$, we define the \emph{class entropy of a filter} to be the entropy of the normalized histogram of class labels that have an associated score $\geq \tau$.
A low class entropy means that at scores above $\tau$, the filter is very class selective.
As this threshold changes, the class entropy traces out a curve which we call the \emph{entropy curve}.
The \emph{area under the entropy curve} (AuE), summarizes the class entropy at all thresholds and is used as a measure of discriminative capacity of the filter. 
The lower the AuE value, the more class selective the filter is.
The AuE values are used to sort filters in Section \ref{sub:fine-entropy}.

%We can also characterize the distribution of discriminative ability of all filters of a layer.
%To do this, we sort the filter according to their AuE.
%Then, the cumulative sum of AuE values in the sorted list is calculated (called \emph{cumulative AuE} or CAuE). 
%The $i$-th entry of the CAuE list is the sum of the AuE scores of the top $i$ most discriminative filters.
%The difference in the value of the $i$-th entry before and after fine-tuning measures the change in class selectivity of the top $i$ most discriminative filters due to fine-tuning.
%For comparing results across different layers, the CAuE values are normalized to account for different numbers of filters in each layer. 
%Specifically, the $i$-th entry of the CAuE list is divided by $i$. 
%This normalized CAuE is called the Mean Cumulative Area Under the Entropy Curve (MCAuE).
%A lower value of MCAuE indicates that the set filters is more discriminative.

\bibliographystyle{splncs03}
\bibliography{egbib}

%\bibliographystyle{splncs03}
%\bibliography{egbib}

\clearpage
\setcounter{section}{0}
\setcounter{figure}{0}
\noindent\makebox[\linewidth]{\rule{\linewidth}{3.0pt}}
\begin{center}
\LARGE{\textbf{Supplementary Material}}
\end{center}
\noindent\makebox[\linewidth]{\rule{\linewidth}{0.8pt}}

\section{Effect of fine-tuning on CNN parameters}
In the main paper we provided evidence that fine-tuning a discriminatively pre-trained network is very effective in terms of task performance.
We also provided insights into how fine-tuning changes its parameters.
Here we describe and discuss in greater detail some metrics for determining the effect of fine-tuning.
\subsection{Defining the measure of discriminative capacity of a filter}
\label{sub:fine-entropy}
The entropy of a filter is calculated to measure its discriminative capacity. The use of entropy is motivated by works such as \cite{Breiman}, \cite{AmitGeman}.
For computing the entropy of a filter, we start by collecting filter responses from a set of $N$ images.
Each image, when passed through the CNN produces a $p \times p$ heat map of scores for each filter in a given layer (e.g., $p = 6$ for a conv-5 filter and $p = 1$ for an fc-6 filter).
This heat map is vectorized (\texttt{x(:)} in MATLAB) into a vector of scores of length $p^2$. With each element of this vector we associate the class label of the image. 
Thus, for every image we have a score vector and a label vector of length $p^2$ each.
Next, the score vectors from all $N$ images are concatenated into an $Np^2$-length score vector.
The same is done for the label vectors. We define the entropy of a filter in the following three ways.

\subsubsection{Label Entropy.}
\label{sub:def-label-ent}
For a given score threshold $\tau$, we define the \emph{class entropy of a filter} to be the entropy of the normalized histogram of class labels that have an associated score $\geq \tau$.
A low class entropy means that at scores above $\tau$, the filter is very class selective.
As this threshold changes, the class entropy traces out a curve which we call the \emph{entropy curve}.
The \emph{area under the entropy curve} (AuE), summarizes the class entropy at all thresholds and is used as a measure of discriminative capacity of the filter. 
The lower the AuE value, the more class selective the filter is.

\subsubsection{Weighted Label Entropy.}
\label{sub:def-weighted-label-ent}
While computing the class label histogram, instead of the label count we use the sum of the scores associated with the labels to construct the histogram. (Note: Since we are using outputs of the rectified linear units, all scores are $\geq 0$.)

\subsubsection{Spatial-Max (spMax) Label Entropy.}
\label{sub:def-spmax-label-ent}
Instead of vectorizing the heatmap, the filter response obtained as a result of max pooling the $p \times p$ filter output is associated with the class label of each image. Thus, for every image we have a score vector and a class label vector of length $1$ each.
Next, the score vectors from all $N$ images are concatenated into an $N$-length score vector.
Then, we proceed in the same way as for the case of Label Entropy to compute the AuE of each filter. 

\subsection{Defining the measure of discriminative capacity of a layer}
The discriminative capacity of layer is computed as following: The filters are sorted in increasing order of their AuE.
Next, the cumulative sum of AuE values in this sorted list is calculated. The obtained list of Cumulative AuEs is referred to as CAuE. 
Note that, the $i$-th entry of the CAuE list is the sum of the AuE scores of the top $i$ most discriminative filters.
The difference in the value of the $i$-th entry before and after fine-tuning measures the change in class selectivity of the top $i$ most discriminative filters due to fine-tuning.
For comparing results across different layers, the CAuE values are normalized to account for different numbers of filters in each layer. 
Specifically, the $i$-th entry of the CAuE list is divided by $i$. 
This normalized CAuE is called the Mean Cumulative Area Under the Entropy Curve (MCAuE).
A lower value of MCAuE indicates that the individual filters of the layer are more discriminative.

\setlength{\tabcolsep}{1pt}
\begin{table}
\begin{center}
\caption{This table lists percentage decrease in MCAuE as a result of finetuning when only 0.1, 0.25, 0.50 and 1.00 fraction of all the filters were used for computing MCAuE. A lower MCAuE indicates that filters in a layer are more selective/class specific. The 0.1 fraction includes the top 10\% most selective filters, 0.25 is top 25\% of most selective filters. Consequently, comparing MCAuE at different fraction of filters gives a better sense of how selective the ``most" selective filters have become.
A negative value in the table below indicates increase in entropy. Note that for all the metrics maximum decrease in entropy takes place while moving from layer 5 to layer 7. Also, note that for fc-6 and fc-7 the values in Label Entropy and spMax Label Entropy are same as these layers have spatial maps of size 1.}
\label{table:fine-change}
\scalebox{0.85}{
\newcolumntype{d}[2]{D{.}{\cdot}{#1} }
\begin{tabular}{|c|p{1cm} p{1cm} p{1cm} p{1cm}| p{1cm} p{1cm} p{1cm} p{1cm}|p{1cm} p{1cm} p{1cm} p{1cm}|}
\hline
Layer & \multicolumn{4}{c|}{Label Entropy} & \multicolumn{4}{c|}{Weighted Label Entropy} & \multicolumn{4}{c|}{spMax Label Entropy} \\
\hline
& 0.1 & 0.25 & 0.5 & 1.0 & 0.1 & 0.25 & 0.5 & 1 & 0.1 & 0.25 & 0.5 & 1.0 \\
\hline
conv-1 & $-0.02$  & $-0.14 $  & $-0.19$  & $-0.19$  & $0.06$  & $-0.13$  & $-0.16$  & $-0.16$  & $0.19$  & $0.10$  & $0.07$  & $0.04$   \\
conv-2 & $-0.71$  & $-0.31$   & $-0.14$  & $0.01$  & $0.41$  & $0.53$  & $0.58$  & $0.57$  & $-0.39$  & $-0.03$  & $0.11$  & $0.23$  \\ 
conv-3 & $-1.14$  & $-0.86$   & $-0.67$  & $-0.44$ & $1.11$  & $0.66$  & $0.52$  & $0.32$ & $0.14$  & $0.20$  & $0.32$  & $0.33$   \\
conv-4 & $-0.54$  & $-0.31$  & $-0.19$  & $-0.05$ & $-0.10$  & $0.55$  & $0.64$  & $0.57$  & $0.93$  & $0.97$  & $0.80$  & $0.65$ \\
conv-5 & $0.97$  & $0.55$  & $0.43$  & $0.36$ & $5.84$  & $3.53$  & $2.66$  & $1.85$  & $4.87$  & $3.05$  & $2.31$  & $1.62$   \\
fc-6 & $6.52$  & $5.06$  & $3.92$  & $2.64$  & $9.59$  & $7.55$  & $6.08$  & $4.27$ & $6.52$  & $5.06$  & $3.92$  & $2.64$   \\
fc-7 & $5.17$  & $2.66$  & $1.33$  & $0.44$  & $20.58$  & $14.75$  & $11.12$  & $7.78$  & $5.17$  & $2.66$  & $1.33$  & $0.44$  \\
\hline
\end{tabular}}
\end{center}
\end{table}
\setlength{\tabcolsep}{1.4pt}

\begin{figure}[t!]
\centering
\subfloat[Weighted Label Entropy]{\includegraphics[width=0.95\linewidth]{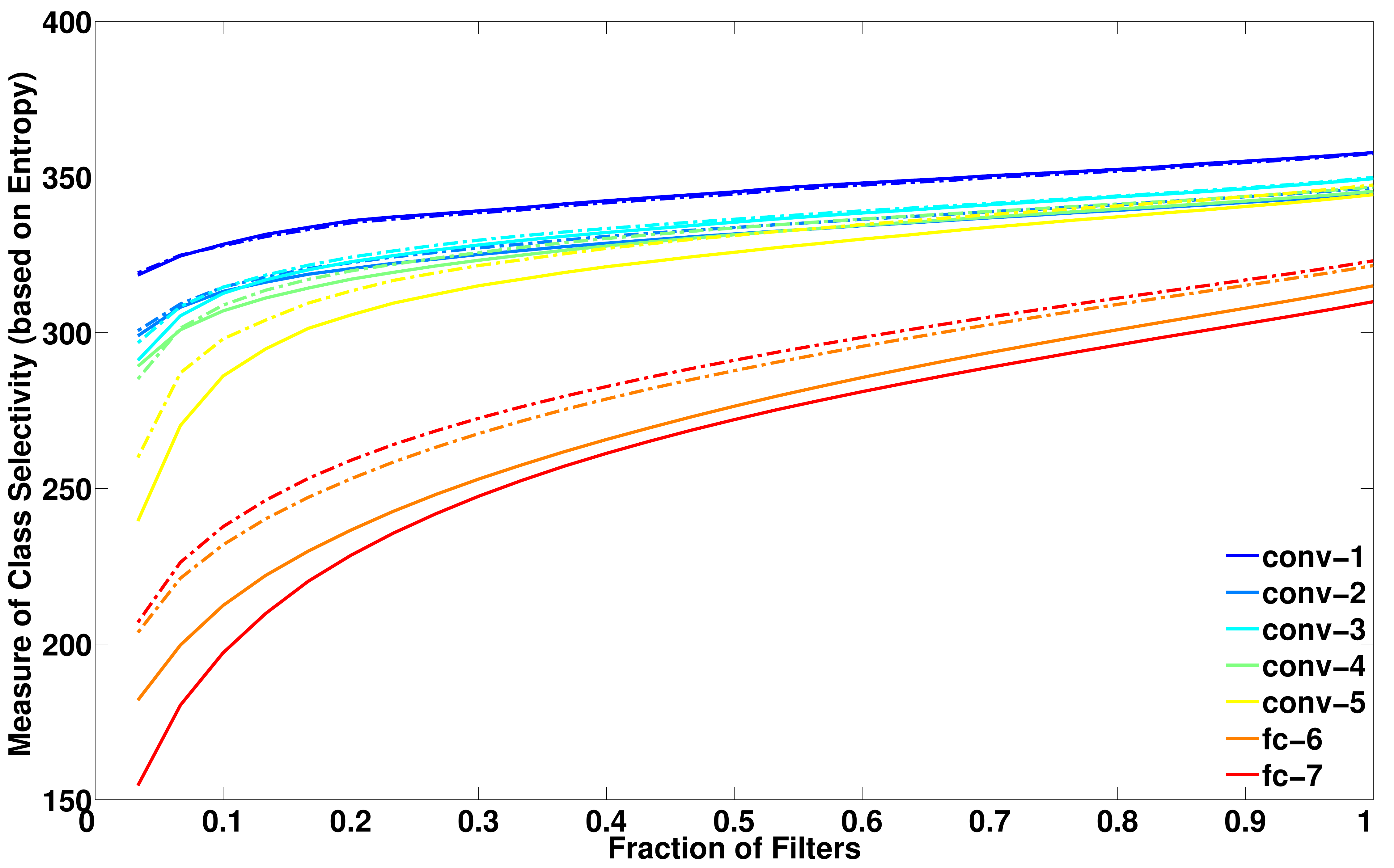}} \\
\subfloat[Spatial-Max Label Entropy]{\includegraphics[width=0.95\linewidth]{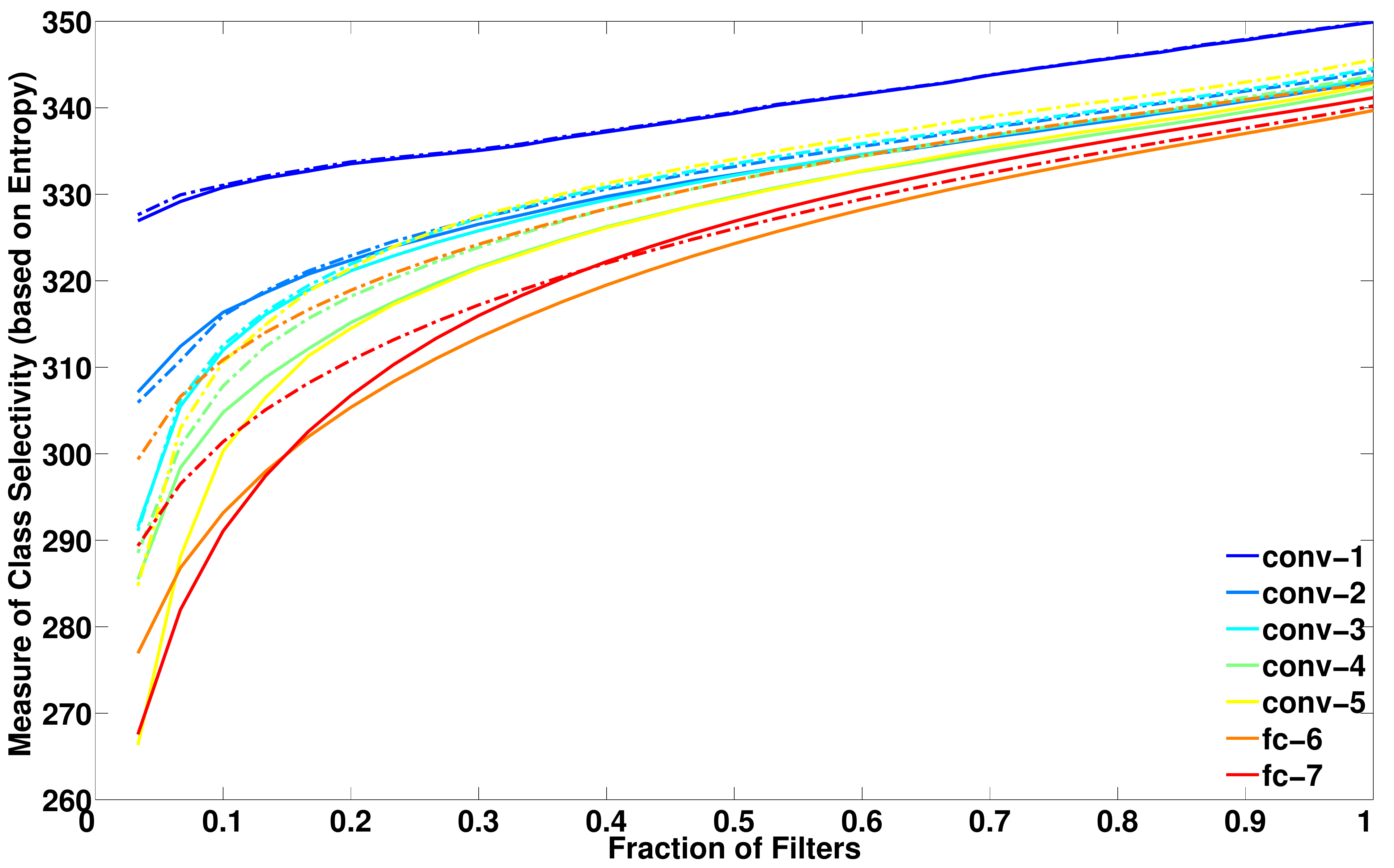}}
\caption{PASCAL object class selectivity (measured as MCAuE) plotted against the fraction of filters, for each layer, before fine-tuning (dash-dot line) and after fine-tuning (solid line). A lower value indicates greater class selectivity. (a),(b) show MCAUE computed using Weighted-Label-Entropy and Spatial-Max Label-Entropy method respectively.}
\label{fig:fine-entropy}
\end{figure}

\subsection{Discussion}
The MCAuE measure of determining layer selectivity before and after fine-tuning is shown in Figure \ref{fig:fine-entropy} for Weighted Label and Spatial-Max Label Entropy. Results for Label Entropy method are presented in the main paper. A quantitative measure of change in entropy due to finetuning, computed as percentage change is defined as following:
\begin{eqnarray}
\text{Percent Decrease} = 100 \times \frac{MCAuE_{pre} - MCAuE_{fine}}{MCAuE_{pre}}
\end{eqnarray}
where, $MCAuE_{fine}$ is for fine-tuned network and $MCAuE_{untuned}$ is for network trained on imagenet only. The results are summarized in table \ref{table:fine-change}.
 
As measured by Label Entropy, layers 1 to 5 undergo negligible change in their discriminative capacity, whereas layers 6-7 become a lot more discriminative. Whereas, the measures of Weighted Label and Spatial-Max Label Entropy indicate that only layers 1 to 4 undergo minimal changes and other layers become substantially more discriminative. These results confirm the intuition that lower layers of the CNN are more generic features, whereas fine-tuning mostly effects the top layers. Also, note that these results are true for fine-tuning for moderate amount of training data available as part of PASCAL-DET. It is yet to be determined how lower convolutional layers would change due to fine-tuning when more training data is available. 

\section{Are there grandmother cells in CNNs?}
In the main paper we studied the nature of representations in mid-level CNN representations given by conv-5. Here, we address the same question for layer fc-7, which is the last layer of CNN and features extracted from this lead to best performance. The results for number of filters required to achieve the same performance as all the filters taken together is presented in Figure \ref{fig:svm-sel-dims}. Table \ref{table:num-fil} reports the number of filters required per class to obtain 50\% and 90\% of the complete performance. It can be seen that like conv-5, feature representations in fc-7 are also distributed for a large number of classes. It is interesting to note, that for most classes 50\% performance can be reached using a single filter, but for reaching 90\% performance a lot more filters are required. 
\setlength{\unitlength}{\linewidth}
\begin{figure}[t!]
\centering
\begin{picture}(0.04,0.3)(0,0)
\put(0.0,0.1){\rotatebox{90}{\scriptsize{\textbf{Fraction of complete performance}}}}
\end{picture}
\includegraphics[width=0.93\linewidth]{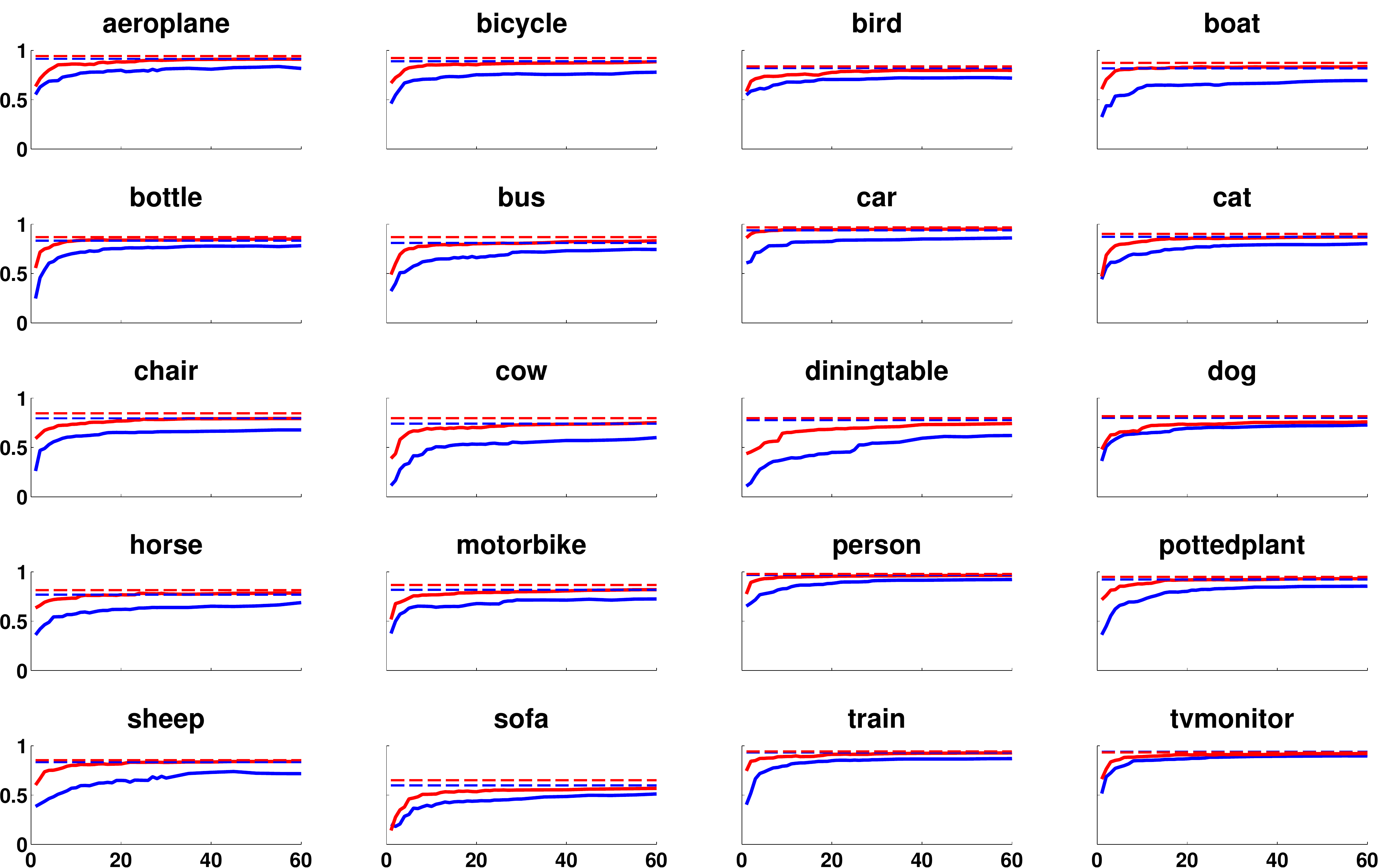}
\begin{picture}(1.0,0.01)(0,0)
\put(0.40,0.0){{\scriptsize{\textbf{Number of fc-7 filters}}}}
\end{picture}
\caption{The fraction of complete performance on PASCAL-DET-GT achieved by fc-7 filter subsets of different sizes. Complete performance is the AP computed by considering responses of all the filters. The solid blue and red lines are for pre-trained and fine-tuned network respectively. The dashed blue and red lines show the complete performance.  Notice, that for a few classes such as person, bicycle and cars only a few filters are required, but for many classes substantially more filters are needed, indicating a distributed code.}
\label{fig:svm-sel-dims}
\end{figure}  

\setlength{\tabcolsep}{1.1pt}
\begin{table}[t!]
\renewcommand{\arraystretch}{1.2}
\begin{center}
\caption{Number of filters required to achieve 50\% or 90\% of the complete performance on PASCAL-DET-GT using a CNN pre-trained on ImageNet and fine-tuned for PASCAL-DET using fc-7 features.}
\label{table:num-fil}
\resizebox{\linewidth}{!}{
\begin{tabular}{l|c||cccccccccccccccccccc}
\noalign{\smallskip}
& perf. & aero & bike & bird & boat & bottle & bus & car & cat & chair & cow & table & dog & horse & mbike & person & plant & sheep & sofa & train & tv \\
\hline
pre-train & 50\% & 1 & 1 & 1 & 2 & 2 & 3 & 1 & 1 & 2 & 6 & 11 & 2 & 2 & 2 & 1 & 3 & 3 & 5 & 2 & 1 \\
fine-tune & 50\% & 1 & 1 & 1 & 1 & 1 & 1 & 1 & 1 & 1 & 2 & 1 & 1 & 1 & 1 & 1 & 1 & 1 & 3 & 1 & 1 \\
\hline
pre-train & 90\% & 33 & 40 & 40 & 40 & 17 & 32 & 32 & 31 & 40 & 40 & 40 & 35 & 37 & 37 & 17 & 29 & 40 & 40 & 17 & 8\\
fine-tune & 90\% & 6 & 7 & 11 & 4 & 5 & 10 & 2 & 8 & 19 & 27 & 32 & 18 & 9 & 16 & 2 & 7 & 7 & 40 & 3 & 4 
\end{tabular}
}
\end{center}
\end{table}
\setlength{\tabcolsep}{1.4pt}

%\section{Conclusion}
%\label{sec:conclusion}
%In this paper we analysed different properties of convolutional neural networks with the aim of gaining insights required to efficiently exploit the rich feature hierarchies provided by such networks. \\

%\bibliographystyle{splncs03}
%\bibliography{egbib}

\end{document}